\documentclass{article}

    \usepackage[preprint, nonatbib]{neurips_2019}

\usepackage[utf8]{inputenc} 
\usepackage[T1]{fontenc}    
\usepackage{hyperref}       
\usepackage{url}            
\usepackage{booktabs}       
\usepackage{amsfonts}       
\usepackage{nicefrac}       
\usepackage{microtype}      

\usepackage{amsmath}
\usepackage{graphicx}
\usepackage{subfigure}
\usepackage{diagbox}
\usepackage{multirow}
\usepackage{caption}
\usepackage{wrapfig}
\usepackage{color}

\graphicspath{{figure/}}

\title{Challenge of Spatial Cognition for Deep Learning}

\author{%
  Xi Zhang\\
  Dept. of Electronic Engineering\\
  Shanghai Jiao Tong University\\
  \texttt{zhangxi\_19930818@sjtu.edu.cn}\\
  \And
  Xiaolin Wu\\
  Dept. of Electrical \& Computer Engineering\\
  McMaster University\\
  \texttt{xwu@ece.mcmaster.ca}\\
  \And
  Jun Du\\
  Dept. of Electronic Engineering\\
  Shanghai Jiao Tong University\\
  \texttt{dujunandy0113@sjtu.edu.cn} \\
}

\begin{document}

\maketitle

\vskip -0.4cm
\begin{abstract}
Given the success of the deep convolutional neural networks (DCNNs) in applications of visual recognition and classification, it would be tantalizing to test if DCNNs can also learn spatial concepts, such as straightness, convexity, left/right, front/back, relative size, aspect ratio, polygons, etc., from varied visual examples of these concepts that are simple and yet vital for spatial reasoning.
Much to our dismay, extensive experiments of the type of cognitive psychology demonstrate that the data-driven deep learning (DL) cannot see through superficial variations in visual representations and grasp the spatial concept in abstraction.
The root cause of failure turns out to be the learning methodology, not the computational model of the neural network itself.
By incorporating task-specific convolutional kernels, we are able to construct DCNNs for spatial cognition tasks that can generalize to input images not drawn from the same distribution of the training set.  This work raises a precaution that without manually-incorporated priors or features DCCNs may fail spatial cognitive tasks at rudimentary level.
\end{abstract}

\section{Introduction}
The past decade has witnessed rapid advances of deep learning as a powerful problem-solving paradigm, with applications in almost all fields of engineering, sciences and medicine.  In particular, deep convolutional neural networks (DCNNs) are lauded for their apparent visual intelligence, by which we refer to the successes acclaimed by DCNNs in visual pattern analysis, recognition and classification tasks
\cite{AlexNet,VGG,GoogleNet,ResNet,GAN,Tang2014,FaceNet,Deepid3,DenseNet}.
On the other hand, deep learning still falls short of the strict definition of artificial intelligence, as many academics contend, because it cannot match the cognition and abstraction power of human brain.  Indeed, all the successes of the DCNN
learning approach in visual computing fall into the type of statistical inferences under the i.i.d. condition, a problem setting that practitioners take for granted and are contented with.  Considering that for humans the most primitive cognitions are vision-related space
awareness and reasoning, it will be interesting and tantalizing to examine whether DCNNs can pass any Turing tests of spatial cognition type.

In order to contrast the abilities of various DCNNs in spatial cognition against that of humans, we design and subject deep learning to a family of experiments similar to those in cognitive psychology, and evaluate how well pure data-driven DCNNs can, under various levels of supervision, learn a simple spatial concept, such as left/right, front/back, relative size, aspect ratio, straightness, convexity, polygons, etc., from a training set of visual depictions of the pertaining concept, very much like picture cards used to teach children.  The training images for a specific spatial relationship/property are made highly diversified at the visual signal level;  the objects in these training images vary in the color, size, shape, orientation and position.
The number of examples to teach DCNNs a spatial concept is sufficiently large but by no means exhaustive in the sample space.  The main purpose of our experiments is to examine how well widely used DCNNs, including AlexNet, VGGNet, ResNet and DenseNet, can generalize beyond the i.i.d. limitations on spatial cognition tasks of different difficulties and representation complexities, i.e., see through superficial variations of visual signals and arrive at an abstract conceptualization.

This work is the first systematical study of DCNNs' potentials in basic spatial cognition and reasoning tasks.  Our findings should be of interests and significance to researchers who apply deep learning methods to solve vision problems, as many applications of visual intelligence (e.g., path planning and scene understanding) henge on understanding of and reasoning with spatial properties of objects and events.
A similar study on the limitations of machine intelligence but in the field of natural language processing, is the Winograd Schema Challenge (WSC) posed by Hector Levesque~\cite{wsc,evan2014,terry1962}.  The WSC test includes many questions that target qualitative spatial distinctions.  The difference between our work and WSC is that the latter designs  cognitive questions via natural language conversations, whereas our questions are posed in visual representations.

Besides the relevance to applied machine vision research, we choose spatial cognition as a telling case study to assess the limitations of DCNNs in cognitive computing, due to the primitivity and necessity of space awareness and reasoning for daily functioning of humans and other animals.  The type of cognitive tasks used by this paper to challenge DCNNs can be performed by small children with ease, accuracy and confidence.  Indeed, spatial cognition capability is innate for infants, as supported by the discovery of place cells in the hippocampus and the role of the hippocampus in spatial cognition by Nobel laureate
O'Keefe and his coauthors \cite{o1998place,book1978}.



%





In a similar approach, Ritter et al.~\cite{Ritter} studied the interpretability of DCNNs using the methods of cognitive psychology. Their findings are very interesting: one shot learning methods trained on ImageNet exhibit a human-like bias when characterizing a class of objects with a word or label. Encouraged by their experimental results, the authors promote to leverage tools of cognitive psychology in the research of interpreting the behavior of DCNNs.

Very recently, Wu et al.~\cite{Wu2018} conducted Turing-type tests of data-driven DCNNs on the cognitive task of numerosity.  Their findings are largely negative.  Deep learning was found incapable to extract the abstract notion of natural numbers from a training set of rich visual representations of numbers. Considering that the number perception is a cognitive ability innate to human and primates \cite{Reas,Harvey,Brannon,Burr,Dehaene,Nieder,Xu}, this recent study reveals a quite severe cognitive deficit of deep learning.

\section{Key Findings}

Along the above line of enquiry, this new study probes the boundary of cognitive limitations of deep learning further inwards. This is because, at least spatial relationships and properties of objects and events, such as left/right, straightness, etc., are more intuitive and primitive than distilling the abstract notion of numbers from concrete examples of 3 apples and 2 cars, etc.  We design and conduct three groups of spatial cognitive experiments for DCNNs: relative positioning, relative size and shape.  Four well-known DCNN architectures AlexNet, VGGNet, ResNet and DenseNet are tested, and their performances are compared.  To evaluate the generalization or abstraction abilities of the DCNNs, we represent a same spatial relationship or property by two versions of sample images, one version of two colors and the other of three colors.  This allows us to examine the cognition power of the DCNNs in relation to the 2D signal complexity.

Much to our disappointment, even on some very simple spatial cognitive tasks, the performances of data-driven DCNNs still do not meet the expectations for “intelligent” machines.  For instance, on the task of comparing the sizes of two objects (the simplest metric property), the size discriminators constructed by deep learning fail to generalize beyond the sample distribution of the training set.  We analyze the failures of the four DCNNs in our spatial cognitive experiments and identify their typical problematic behavior. They are easily diverted by immaterial variations in visual representations of a given spatial property or concept and fail to discover the invariant of the training images that is the pertaining abstraction, whereas preschoolers can perform such spatial judgement and reasoning effortlessly.

In our experiments we also try to measure the performance deteriorations of the four DCNNs in spatial cognition tasks as the visual signal complexity and the inference difficulty increase.  The spatial cognition accuracy of the DCNNs drops by around 15\% if the number of colors used to depict the learning problem increases from two to three.  We test the generalization capabilities of the four DCNNs over varying object size, and carry out
two types of inferences, interpolation-type and extrapolation-type.
In the interpolation-type inference experiments, the training images contain objects of extreme size, either small or large, whereas the objects in the test images have a size in the intermediate range; the extrapolation-type inference experiments are vice versa.
The spatial cognition performances of the four DCNNs for interpolation-type inferences are considerably better than for extrapolation-type inferences, by as much as 20\% in some cases.

%
%

The exposed deficiency of deep learning in spatial cognition is not due to the connectionist computation model of neural networks itself, but to the limitation of the backpropagation learning methodology. By incorporating task-specific convolutional kernels, we are able to construct DCNNs for spatial cognition tasks that can generalize to input images not drawn from the same distribution of the training set. These hand-crafted DCNNs can acquire, via supervised learning, the ability of spatial cognition with a generalization strength exceeding that of the i.i.d. statistical inference. We emphasize that the incorporation of human insight into the neural network is necessary for the quantum jump from pattern matching to spatial cognition; otherwise, the pure data-driven DCNNs cannot achieve the same level of generalization.


\section{Methods and Detailed Results}

In this section we present the design, justifications, and results of our experiments to evaluate the capability of different DCNNs to perform three different types of spatial cognitive tasks.


\subsection{Relative positioning}

Relative positioning of two objects in space is of the simplest type of visual cognition ability for human and other animals. We design and subject DCNNs to cognitive experiments for relative spatial positioning, and examine whether the black box deep learning can comprehend spatial relationships as basic as left versus right and front versus back.

\subsubsection{Left and right}
\begin{figure}[!ht]
\begin{minipage}{0.3\linewidth}
\setlength{\fboxsep}{0pt}
  \centering
    \includegraphics[width=0.3\linewidth]{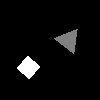}
    \fbox{\includegraphics[width=0.3\linewidth]{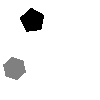}}
    \includegraphics[width=0.3\linewidth]{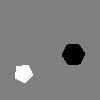}
    \\
    \includegraphics[width=0.3\linewidth]{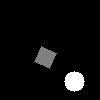}
    \fbox{\includegraphics[width=0.3\linewidth]{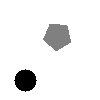}}
    \includegraphics[width=0.3\linewidth]{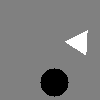}
  \caption{Training images for left vs. right concept.}
  \label{fig:left/right}
\end{minipage}
\hskip 0.04\linewidth
\begin{minipage}{0.3\linewidth}
\setlength{\fboxsep}{0pt}
  \centering
    \includegraphics[width=0.3\linewidth]{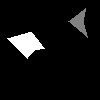}
    \fbox{\includegraphics[width=0.3\linewidth]{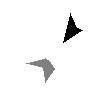}}
    \includegraphics[width=0.3\linewidth]{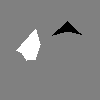}
  \\
    \includegraphics[width=0.3\linewidth]{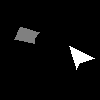}
    \fbox{\includegraphics[width=0.3\linewidth]{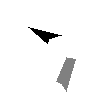}}
    \includegraphics[width=0.3\linewidth]{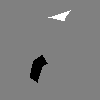}
  \caption{Test images of non-convex polygons.}
  \label{fig:left/right-nc}
\end{minipage}
\hskip 0.04\linewidth
\begin{minipage}{0.3\linewidth}
\setlength{\fboxsep}{0pt}
  \centering
    \includegraphics[width=0.3\linewidth]{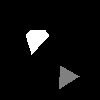}
    \fbox{\includegraphics[width=0.3\linewidth]{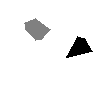}}
    \includegraphics[width=0.3\linewidth]{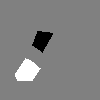}
  \\
    \includegraphics[width=0.3\linewidth]{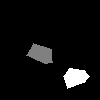}
    \fbox{\includegraphics[width=0.3\linewidth]{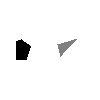}}
    \includegraphics[width=0.3\linewidth]{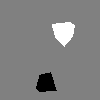}
  \caption{Test images of irregular convex polygons.}
  \label{fig:left/right-ir}
  \end{minipage}
\end{figure}

In our experiments, we use varied visual representations of a simple spatial relationship to train DCNNs to cognize the relationship.  For example, for learning the notions of left and right, we use a set of training images as illustrated in Fig.~\ref{fig:left/right}.  These training images have three colors: black, grey, and white.  The two objects in question and the background are each assigned to a different color at random with equal probability.  The objects are randomly placed and have different shapes ($n$-gons, n=3,4,5,6, and circles) and sizes.  The cognition of relative spatial positions is modeled as a binary classification problem for DCNNs, namely, whether the brighter object is to the left or right of the darker object. All training images are so labeled and used to train a DCCN classifier.  Within a given class, beneath all the superficial variations in color, size and shape, there is an invariant that is the relative position of the two objects. This invariant is the essence to distill by the DCNN learner.

We choose AlexNet as the DCNN for learning the left and right concept, and train it as a binary classifier, outputting 1 if the brighter object is to the left of the darker object, 0 otherwise.
A total of 2400 images (1200 per class) are used for training.
The trained left-right DCNN classifier achieves a perfect 100\% classification rate on the test images that are independently identically distributed as the training images.

To verify whether the left-right DCNN classifier can generalize outside of the distribution of the training images, we test it on test images in which the object size exceeds the range of those in the training images. The test results are tabulated in Table~\ref{tab:left/right}.

We also try to generalize the left-right DCNN to differently shaped objects. Note that all polygons in training images are convex and regular, we test the left-right DCNN with test images consisting of non-convex polygons (Fig.~\ref{fig:left/right-nc}) and irregular convex polygons (Fig.~\ref{fig:left/right-ir}). The performance results are listed in Table~\ref{tab:left/right}.

\begin{table}[!ht]
\centering
  \captionof{table}{Generalization performances of the AlexNet left-right classifier.}
  \label{tab:left/right}
  \begin{tabular}{ccccc}
    \toprule
    objects scaled up 50\% & objects scaled down 50\% & irregular convex & non-convex \\
    \midrule
    100\% & 96.1\% & 99.5\% & 100\% \\
    \bottomrule
  \end{tabular}
\end{table}

\subsubsection{Front and back}
\begin{figure}[!ht]
\centering
\begin{minipage}{0.3\linewidth}
\setlength{\fboxsep}{0pt}
  \centering
    \includegraphics[width=0.3\linewidth]{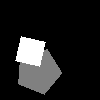}
    \fbox{\includegraphics[width=0.3\linewidth]{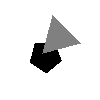}}
    \includegraphics[width=0.3\linewidth]{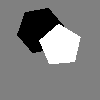}
  \\
    \includegraphics[width=0.3\linewidth]{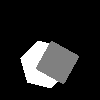}
    \fbox{\includegraphics[width=0.3\linewidth]{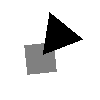}}
    \includegraphics[width=0.3\linewidth]{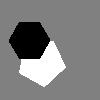}
  \caption{Training images for front vs. back concept.}
  \label{fig:front/back}
\end{minipage}
\hskip 0.2\linewidth
\begin{minipage}{0.3\linewidth}
\setlength{\fboxsep}{0pt}
  \centering
    \includegraphics[width=0.3\linewidth]{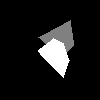}
    \fbox{\includegraphics[width=0.3\linewidth]{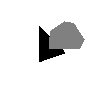}}
    \includegraphics[width=0.3\linewidth]{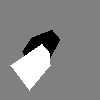}
  \\
    \includegraphics[width=0.3\linewidth]{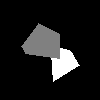}
    \fbox{\includegraphics[width=0.3\linewidth]{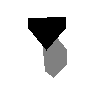}}
    \includegraphics[width=0.3\linewidth]{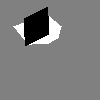}
  \caption{Test images of irregular convex polygons.}
  \label{fig:front/back-shape}
\end{minipage}
\end{figure}

The above experiments show that the black box deep learning can, with supervision, extract the general notions of left and right from non-enumerative example images.  But is this only an isolated case, or the above success of deep learning can be extended to other spatial cognition problems?  To explore further, we next test DCCNs on a similar spatial cognition task: judging which of the two objects is in front of the other.  The training images to represent the front-back relationship are shown in Fig.~\ref{fig:front/back}. These training images for the front-back cognition are of the same design as in Fig.~\ref{fig:left/right}, with the same variations in color, size and shape; the only difference is that the visually encoded relative spatial position is front or back rather than left or right.  Note that all objects in the training images are convex. As such, there is no ambiguity as to which object is in front for the occluded object will have a concave exposed part.

The front-back cognition problem can be modeled as one of binary classification, outputting 1 if the brighter object is in front of the darker object or 0 vice versa.
To test the state of the art DCNNs, we train four DCNNs AlexNet, VGGNet, ResNet and DenseNet for the font-back classification.
A total of 9600 training images (4800 per class) are used to train the four DCNN front-back classifiers.
Under the i.i.d.~condition the four front-back DCNN classifiers perform very well, achieving an accuracy of almost $100\%$.  This is well expected, but our probe is whether they can also make valid inferences of the spatial relationship on test images outside of the distribution of the training images.  To answer the question, we test the four front-back DCNN classifiers on images of objects whose size is not in the range of those in the training set; the inference is required either of the size interpolation or of size extrapolation type.  The performance results for size generalization are listed in Table~\ref{tab:front/back}.  The inference accuracy for size extrapolation is worse than for size interpolation, decreasing 10\% or more; here the AlexNet is the winner.  However, in the case of size interpolation, except the AlexNet the other three DCNNs generalize quite well to object size.

In the training images all objects are regular polygons and circles.  The last row of Table ~\ref{tab:front/back} shows how well the four front-back DCNN classifiers generalize to objects of irregular but still convex polygons (see Fig.~\ref{fig:front/back-shape}).  Although the object shapes in Fig.~\ref{fig:front/back-shape} only differ slightly from those in Fig.~\ref{fig:front/back}, the classification precision of the front-back DCNN drops appreciably.  Here DenseNet-121 performs the best.

\begin{table}[!ht]
\centering
  \captionof{table}{Generalization performances of the four front-back DCNN classifiers.}
  \label{tab:front/back}
  \begin{tabular}{ccccc}
    \toprule
      & AlexNet & VGG-19 & ResNet-34 & DenseNet-121 \\
    \midrule
    size extrapolation &         \textbf{90.8\%} & 90.2\% & 87.0\% & 89.4\% \\
    \midrule
    size interpolation &         90.5\% & \textbf{97.0\%} & 94.1\% & 94.5\% \\
    \midrule
    irregular convex polygons &  91.1\% & 72.5\% & 86.9\% & \textbf{92.6\%} \\
    \bottomrule
  \end{tabular}
\end{table}

%


\subsection{Relative size}
\begin{figure}[!ht]
\begin{minipage}{0.3\linewidth}
\setlength{\fboxsep}{0pt}
  \centering
    \includegraphics[width=0.3\linewidth]{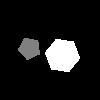}
    \fbox{\includegraphics[width=0.3\linewidth]{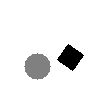}}
    \includegraphics[width=0.3\linewidth]{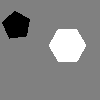}
    \\
    \includegraphics[width=0.3\linewidth]{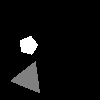}
    \fbox{\includegraphics[width=0.3\linewidth]{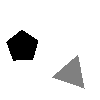}}
    \includegraphics[width=0.3\linewidth]{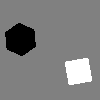}
  \caption{Training images for size cognition.}
  \label{fig:size}
\end{minipage}
\hskip 0.04\linewidth
\begin{minipage}{0.3\linewidth}
\setlength{\fboxsep}{0pt}
  \centering
    \includegraphics[width=0.3\linewidth]{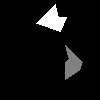}
    \fbox{\includegraphics[width=0.3\linewidth]{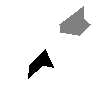}}
    \includegraphics[width=0.3\linewidth]{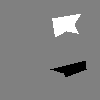}
    \\
    \includegraphics[width=0.3\linewidth]{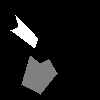}
    \fbox{\includegraphics[width=0.3\linewidth]{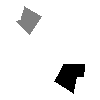}}
    \includegraphics[width=0.3\linewidth]{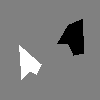}
  \caption{Test images of non-convex polygons.}
  \label{fig:size-nc}
\end{minipage}
\hskip 0.04\linewidth
\begin{minipage}{0.3\linewidth}
\setlength{\fboxsep}{0pt}
  \centering
    \includegraphics[width=0.3\linewidth]{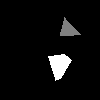}
    \fbox{\includegraphics[width=0.3\linewidth]{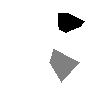}}
    \includegraphics[width=0.3\linewidth]{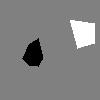}
    \\
    \includegraphics[width=0.3\linewidth]{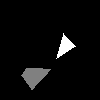}
    \fbox{\includegraphics[width=0.3\linewidth]{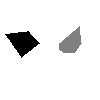}}
    \includegraphics[width=0.3\linewidth]{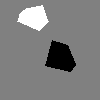}
  \caption{Test images of irregular convex polygons.}
  \label{fig:size-ir}
  \vskip 0.1in
\end{minipage}
\end{figure}

The following experiments are concerned with the assessment of object sizes, the simplest metric property of spatial cognition.  As the visual sense of size needs a reference or unit, we model the cognitive problem as a one of ranking two objects by size, and investigate if it can be solved by a DCNN binary classifier using a set of labeled but non-enumerative training images. The images used to train the size classifier contain two disjoint objects of different sizes (see Fig.~\ref{fig:size}). Like in Fig.~\ref{fig:left/right}, the two objects and background have distinct colors; the objects are regular polygons or circles of random positions and orientations. The only difference is the labeling: 1 if the brighter object is greater than the darker object; 0 otherwise.

Again, we train four DCNN size classifiers using the four well-known network architectures (AlexNet, VGGNet, ResNet, DenseNet).
A total of 2400 (1200 per class) images are used to train the four DCNN size classifiers.
On test images that fall into the same distribution of the training set, the four DCNN size classifiers perform well, nearly flawlessly.
Next, we evaluate the four size classifiers on test images of object size outside of the range of the training set, and check if they can generalize beyond the limitation of the i.i.d. statistical inference.  The performance results of the four DCNN size classifiers for size generalization are tabulated in Table~\ref{tab:size}.  None of the size classifiers can generalize well in the case of size extrapolation; the winner ResNet-34 only achieves an accuracy of 86\%.  In the case of size interpolation, ResNet-34 performs satisfactorily with an accuracy of 98\% while the other three perform quite poorly.

Next we generalize the four size classifiers to slightly different object shapes, from regular convex polygons in the training images to no-convex polygons (Fig.~\ref{fig:size-nc}) and to irregular convex polygons (Fig.~\ref{fig:size-ir}) in the test images, while keeping their sizes within the range of the training set.  These immaterial changes of no effects to human performing the simple cognitive task cause small performance drops for AlexNet, VGG-19 and ResNet-34, but 13\% drop for DenseNet.

\begin{table}[!ht]
\centering
  \captionof{table}{Generalization performances of the four size DCNN classifiers.}
  \label{tab:size}
  \begin{tabular}{ccccc}
    \toprule
      & AlexNet & VGG-19 & ResNet-34 & DenseNet-121 \\
    \midrule
    size extrapolation &        84.4\% & 76.5\% & \textbf{86.4\%} & 73.9\% \\
    \midrule
    size interpolation &        91.6\% & 65.8\% & \textbf{97.8\%} & 88.4\% \\
    \midrule
    irregular convex polygons & 94.6\% & \textbf{95.6\%} & 92.3\% & 87.2\% \\
    \midrule
    non-convex polygons &       92.8\% & \textbf{93.8\%} & 92.2\% & 87.3\% \\
    \bottomrule
  \end{tabular}
\end{table}

\subsection{Shape cognition}
This subsection extends the scope of our inquiry to the ability or limitation of deep learning for shape cognition. Two experiments of shape cognition are conducted: convexity and straightness.
The previous cognition tasks (relative positioning and relative size) need at least three colors to visually depict, but two colors suffice to describe the shape cognition problems here.  In order to examine the generalization capability of the four DCNNs for shape cognition over different signal complexity levels, and compare their performance results with the counterparts for previous spatial cognition problems coded in three colors, we design and conduct shape cognition experiments using both two-color and three-color visual representations of the shape concept.


\subsubsection{Convexity cognition}

\begin{figure}[!ht]
\begin{minipage}{0.35\linewidth}
  \centering
    \includegraphics[width=0.26\linewidth]{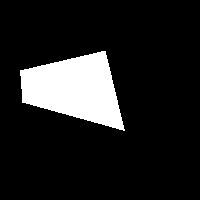}
    \includegraphics[width=0.26\linewidth]{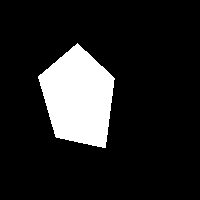}
    \includegraphics[width=0.26\linewidth]{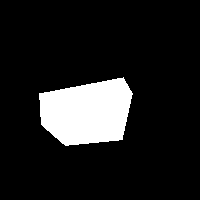}
    \\
    \includegraphics[width=0.26\linewidth]{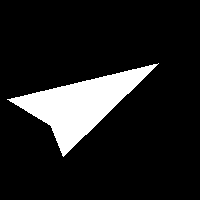}
    \includegraphics[width=0.26\linewidth]{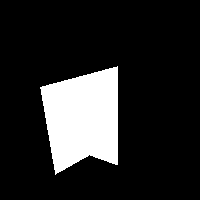}
    \includegraphics[width=0.26\linewidth]{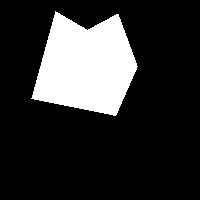}
  \caption{Two-color training images for convexity cognition.}
  \label{fig:convexity-two}
\end{minipage}
\hfill
\begin{minipage}{0.6\linewidth}
\setlength{\fboxsep}{0pt}
  \centering
    \includegraphics[width=0.15\linewidth]{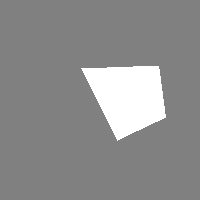}
    \includegraphics[width=0.15\linewidth]{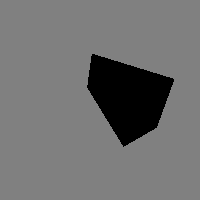}
    \fbox{\includegraphics[width=0.15\linewidth]{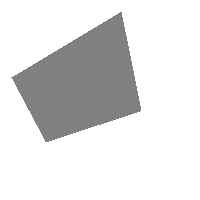}}
    \fbox{\includegraphics[width=0.15\linewidth]{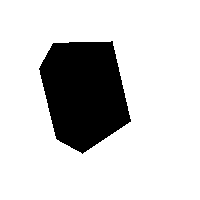}}
    \includegraphics[width=0.15\linewidth]{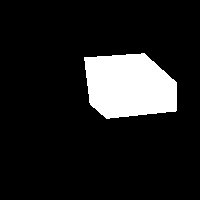}
    \includegraphics[width=0.15\linewidth]{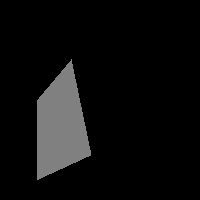}
    \\
    \includegraphics[width=0.15\linewidth]{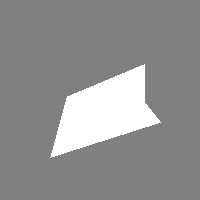}
    \includegraphics[width=0.15\linewidth]{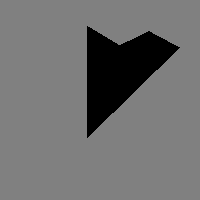}
    \fbox{\includegraphics[width=0.15\linewidth]{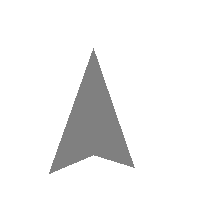}}
    \fbox{\includegraphics[width=0.15\linewidth]{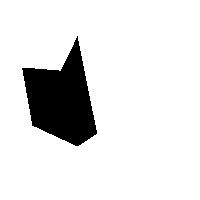}}
    \includegraphics[width=0.15\linewidth]{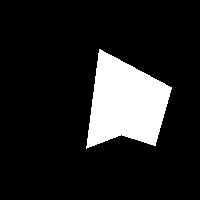}
    \includegraphics[width=0.15\linewidth]{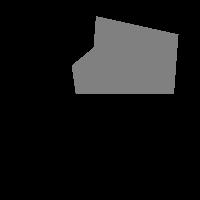}
  \caption{Three-color training images for convexity cognition.}
  \label{fig:convexity-three}
\end{minipage}
\end{figure}

Convexity vs.\ concavity is a key spatial concept as it is a key indicator of many significant events in space; for instance, occluded or broken objects.
The two-color and three-color training images, shown in Fig.~\ref{fig:convexity-two} and Fig.~\ref{fig:convexity-three} respectively, are used to train the four DCNNs.  The objects in the training images are randomly placed and have different shapes ($n$-gons, $n=4,5,6$) and sizes. The training images are labeled 0 or 1 depending on whether the polygons in the images are convex or not.
Like in previous cases, a total of 6000 (3000 per class) training images are used. As expected, these trained networks achieve almost 100\% accuracy on the test images drawn from the same distribution of the training images. 

However, the classification rates of all four DCNNs decrease significantly for the two-color version of the problem, if they are applied to test images containing objects whose size is outside the range of those in the training images, as shown in Table~\ref{tab:convexity-two}.  The results for three-color version of the problem deteriorate further (see Table~\ref{tab:convexity-three}), indicating the negative effect of signal complexity on the generalization.  Also, as in previous experiments, all four DCNNs generalize more poorly in size extrapolation than in size interpolation.  Relatively speaking, VGG-19 and DenseNet-121 perform better than the other two DCNNs for the two-color version.  For the three-color version, practically all four DCNNs fail the inference tests of size extrapolation, whereas only ResNet-34 manages to reach an accuracy of 91\% in the easier case of size interpolation.


\begin{table}[!ht]
\centering
  \vskip -0.4cm
  \captionof{table}{Size generalization of four convexity classifiers for two-color version.}
  \label{tab:convexity-two}
  \begin{tabular}{ccccc}
    \toprule
      & AlexNet & VGG-19 & ResNet-34 & DenseNet-121 \\
    \midrule
    size extrapolation &    68.9\% & 84.9\% & 71.8\% & \textbf{85.8\%} \\
    \midrule
    size interpolation &    81.5\% & \textbf{95.9\%} & 91.2\% & 95.5\% \\
    \bottomrule
  \end{tabular}
\end{table}
\begin{table}[!ht]
\centering
  \vskip -0.5cm
  \captionof{table}{Size generalization of four convexity classifiers for three-color version.}
  \label{tab:convexity-three}
  \begin{tabular}{ccccc}
    \toprule
      & AlexNet & VGG-19 & ResNet-34 & DenseNet-121 \\
    \midrule
    size extrapolation &    61.2\% & 57.2\% & 65.4\% & \textbf{69.2\%} \\
    \midrule
    size interpolation &    75.4\% & 78.1\% & \textbf{90.9\%} & 81.9\% \\
    \bottomrule
  \end{tabular}
\end{table}

\subsubsection{Straightness cognition}
\begin{figure}[!ht]
\begin{minipage}{0.35\linewidth}
  \centering
    \includegraphics[width=0.26\linewidth]{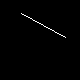}
    \includegraphics[width=0.26\linewidth]{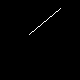}
    \includegraphics[width=0.26\linewidth]{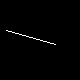}
    \\
    \includegraphics[width=0.26\linewidth]{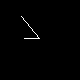}
    \includegraphics[width=0.26\linewidth]{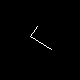}
    \includegraphics[width=0.26\linewidth]{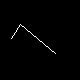}
  \caption{Two-color training images for straightness cognition.}
  \label{fig:lines-two}
\end{minipage}
\hfill
\begin{minipage}{0.6\linewidth}
\setlength{\fboxsep}{0pt}
  \centering
    \includegraphics[width=0.15\linewidth]{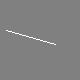}
    \includegraphics[width=0.15\linewidth]{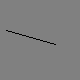}
    \fbox{\includegraphics[width=0.15\linewidth]{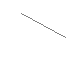}}
    \fbox{\includegraphics[width=0.15\linewidth]{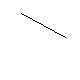}}
    \includegraphics[width=0.15\linewidth]{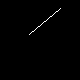}
    \includegraphics[width=0.15\linewidth]{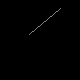}
    \\
    \includegraphics[width=0.15\linewidth]{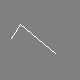}
    \includegraphics[width=0.15\linewidth]{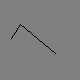}
    \fbox{\includegraphics[width=0.15\linewidth]{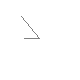}}
    \fbox{\includegraphics[width=0.15\linewidth]{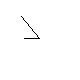}}
    \includegraphics[width=0.15\linewidth]{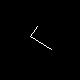}
    \includegraphics[width=0.15\linewidth]{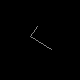}
  \caption{Three-color training images for straightness cognition.}
  \label{fig:lines-three}
\end{minipage}
\end{figure}

One of the simplest spatial percepts is straightness.  The two-color and three-color example images are shown in Fig.~\ref{fig:lines-two} and Fig.~\ref{fig:lines-three}, labeled 1 or 0 depending on whether the line in the image is straight or not.  We model the straightness cognition problem as one of binary classification and train four DCNN binary classifiers for the task with 2400 (1200 per class) training images.  As expected, the four trained DCNN straightness classifiers perform the i.i.d. inference perfectly.  The classification rates for test images drawn from the same distribution of those in Fig.~\ref{fig:lines-two} are nearly perfect $100\%$.  

However, if the length of the line in the test image is outside the range of the training images, then the classification rate drops greatly for the three-color version of the cognition problem (Table~\ref{tab:lines-three}).  Also, it is more difficult to generalize to size extrapolation than to size interpolation.  As in the convexity cognition, all four DCNNs practically fail the inference tests of size extrapolation with three-color test images.   Interestingly, DenseNet-121 performs satisfactorily when being generalized to size interpolation.

Summarizing all experimental results of Tables 1 through 7, there is no clear winner among the four DCNNs for all spatial cognition tasks and in all settings of signal complexity.


\begin{table}[!ht]
\centering
  \vskip -0.4cm
  \captionof{table}{Size generalization of four straightness classifiers for two-color version.}
  \label{tab:lines-two}
  \begin{tabular}{ccccc}
    \toprule
      & AlexNet & VGG-19 & ResNet-34 & DenseNet-121 \\
    \midrule
    size extrapolation &    72.5\% & \textbf{91.2\%} & 82.7\% & 87.3\% \\
    \midrule
    size interpolation &    82.1\% & \textbf{96.6\%} & 95.1\% & 90.4\% \\
    \bottomrule
  \end{tabular}
\end{table}
\begin{table}[!ht]
\centering
  \vskip -0.4cm
  \captionof{table}{Size generalization of four straightness classifiers for three-color version.}
  \label{tab:lines-three}
  \begin{tabular}{ccccc}
    \toprule
      & AlexNet & VGG-19 & ResNet-34 & DenseNet-121 \\
    \midrule
    size extrapolation &    65.8\% & 72.5\% & \textbf{78.4\%} & 75.7\% \\
    \midrule
    size interpolation &    78.8\% & 83.6\% & 80.5\% & \textbf{92.6\%} \\
    \bottomrule
  \end{tabular}
\end{table}

\section{DCNN Solutions Using Hand-crafted Features}

The previous experiments expose the deficiency of pure data-driven deep learning in spatial cognition tasks, which can be performed by children easily.  But this
does not mean that the connectionist DCNN machinery itself is faulty for cognitive computing.  The problem is with the back propagation methodology of DL.  To make this point, we construct a deterministic DCNN for straightness cognition that can handle lines of any length.

As trivially obvious to human observers of our training images, the unique feature that discriminates straight lines from broken ones is the presence of a vertex or corner point.  Although the data-driven DL fails to discover this pivotal feature from training images as demonstrated above, it is natural for algorithm designers to approach the problem by incorporating into the DCNN architecture corner-responsive convolutional kernels.  After these hand-crafted CNN kernels extract the discriminating features from images, the task of straightness cognition becomes simple for the last stage of fully-connected layers.

%

\begin{figure}[!ht]
  \begin{minipage}{0.48\linewidth}
    \centering
    \includegraphics[width=\linewidth, height=0.88\linewidth]{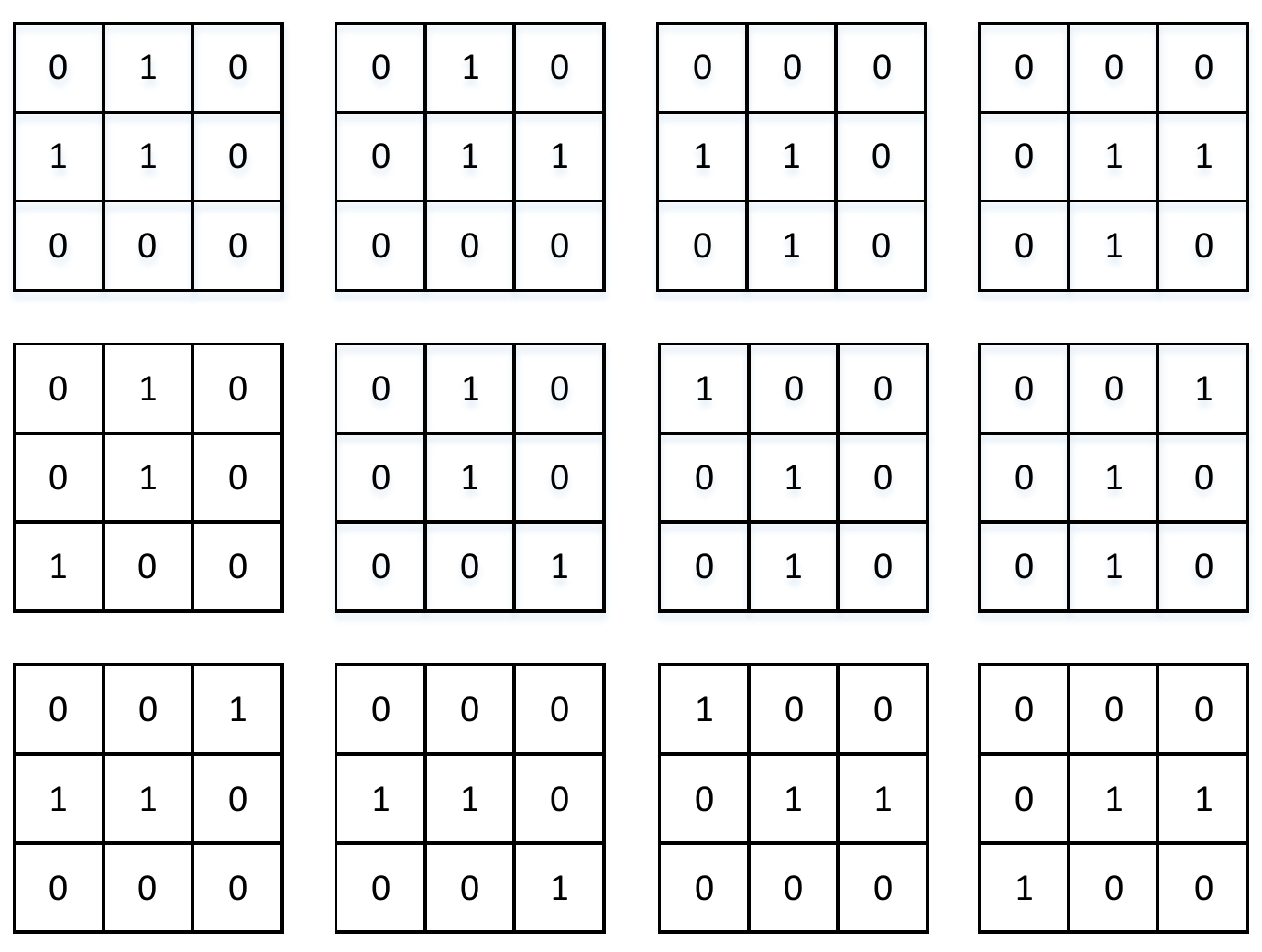}
    \caption{Convolutional kernels of the proposed deterministic DCNN, where 1 and 0 match foreground and background, respectively.}
    \label{fig:kernels}
  \end{minipage}
  \hskip 0.02\linewidth
  \begin{minipage}{0.5\linewidth}
    \centering
    \includegraphics[width=\linewidth, height=0.45\linewidth]{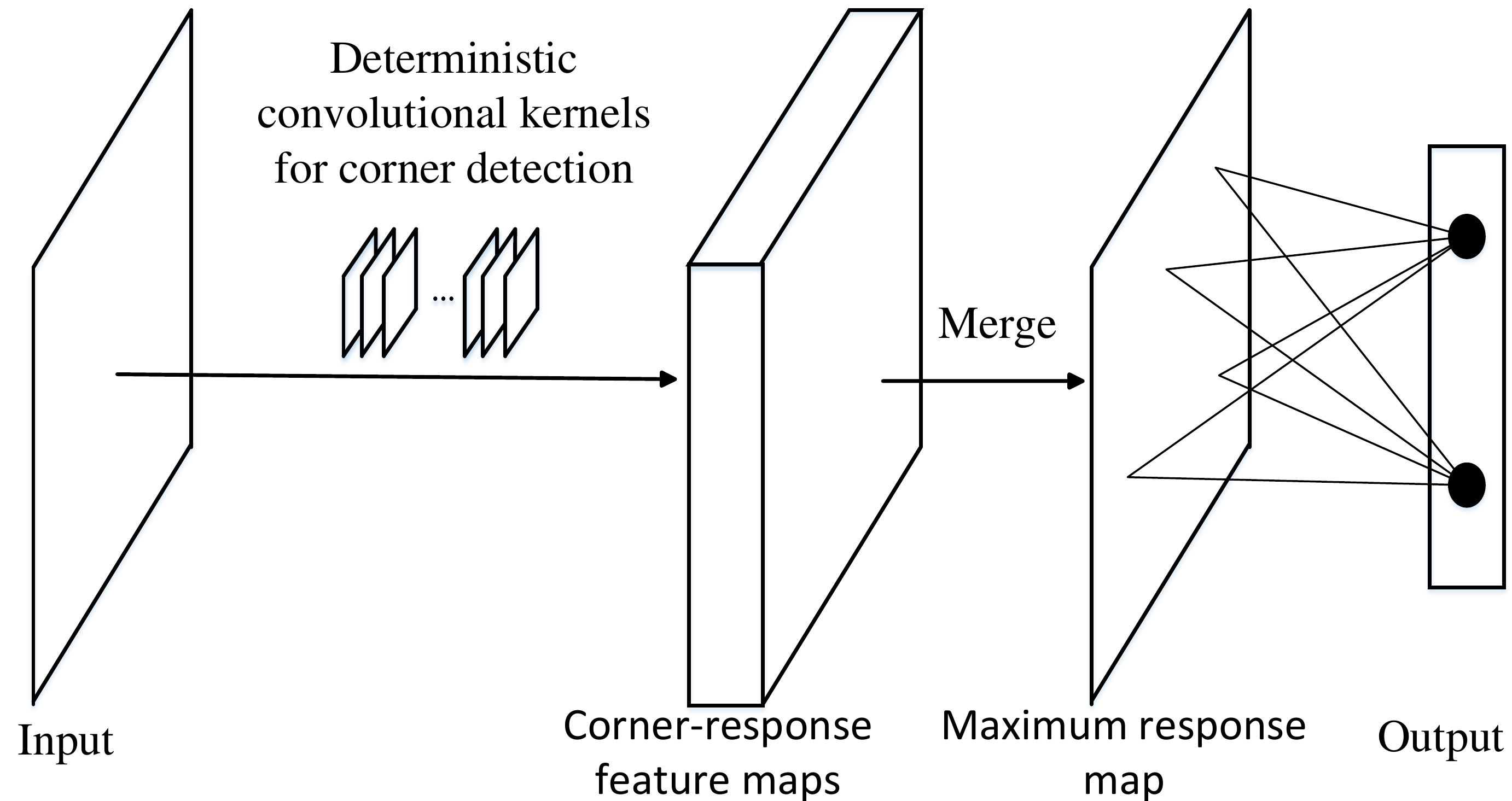}
    \includegraphics[width=\linewidth, height=0.45\linewidth]{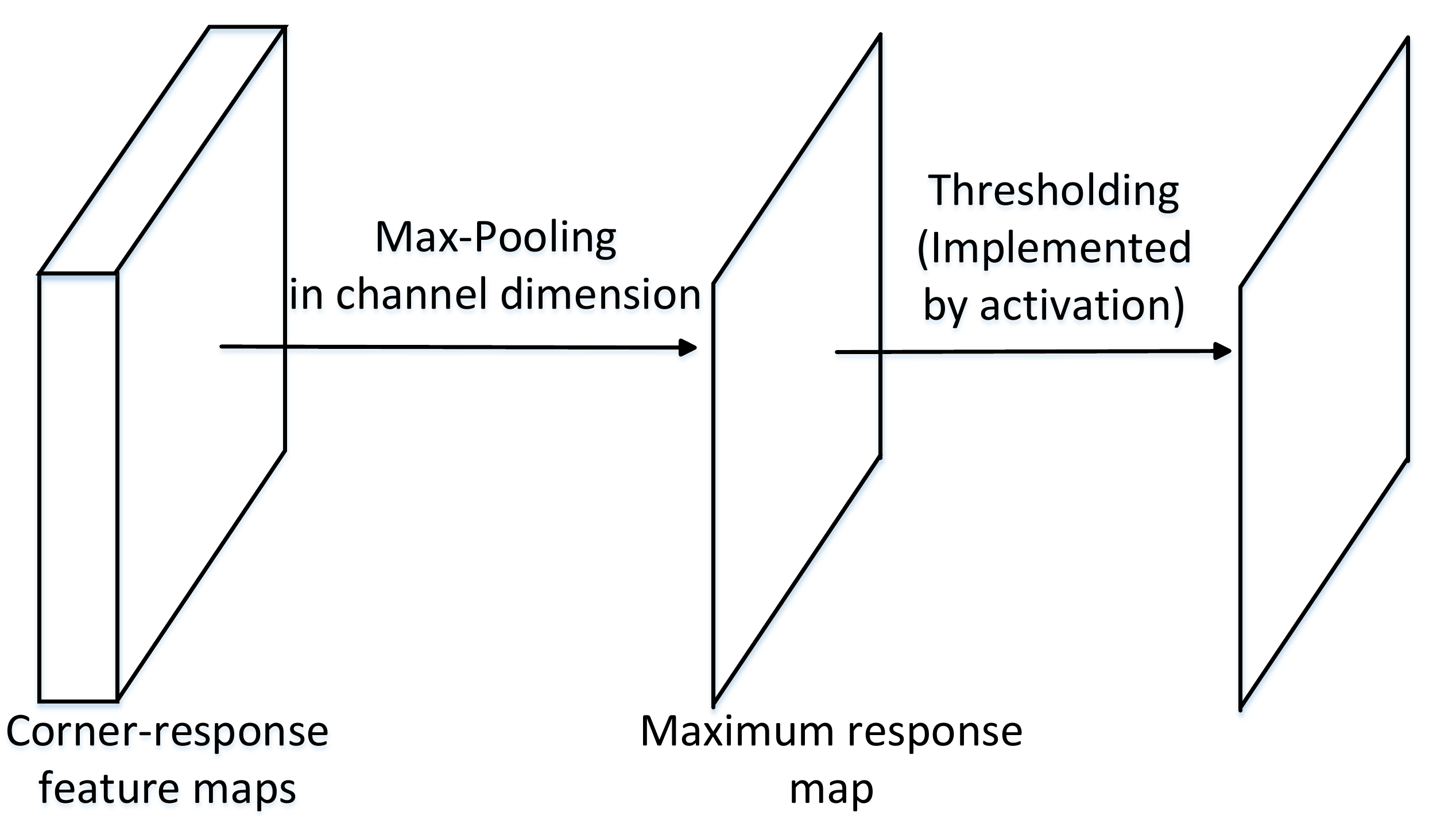}
    \caption{Top: The overall architecture of the proposed deterministic DCNN algorithm.
             Bottom: The detailed merging process.}
    \label{fig:dcnn}
  \end{minipage}
\end{figure}

The said patten matching operation can be carried out by a set of $3\times3$ convolutional kernels (templates).
These convolutional kernels are common corner detection templates in different directions, as illustrated in Fig.~\ref{fig:kernels}.
The detection of the vertices is the key to achieve the generalization in the length of lines, which the black box DL cannot achieve as reported in Table~\ref{tab:lines-two}.  A schematic description of the proposed deterministic DCNN for straightness cognition is given in Fig.~\ref{fig:dcnn}.

In the deterministic DCNN, input image is convolved by the designed convolutional kernels (templates), to generate 12 corner-response feature maps.
These feature maps pass through the max-pooling layer to obtain the maximum response map.
As the responses of straight lines are much smaller than the broken lines to these designed convolutional kernels, we feed the maximum response map into a thresholding layer to  discriminate between straight and broken lines.
The thresholding layer in the neural network is implemented by using a variant of the ReLU activation function.

Despite having only $12\times3\times3=108$ parameters (convolution weights), the
DCNN depicted in Fig.~\ref{fig:dcnn} can learn the spatial concept of straightness using only 2400 training images, and more importantly it can generalize in the line length.  In sharp contrast, the four tested DCNNs, even of millions or more parameters, still fail to truly grasp the simple abstract notion of being straight, confused by length variations.


Next, we examine whether the DL can learn the weights of the twelve convolutional kernels using training images, given the specific network architecture in Fig.~\ref{fig:dcnn}.  This is a much reduced learning problem compared with using a generic DCNN architecture.
We carry out the training of the DCNN for straightness cognition with different random initializations of the weights, and examine if the weights of the twelve $3\times 3$ convolutional kernels in Fig.~\ref{fig:kernels} can be learnt by using the gradient descent method.
The experimental results are disappointing; the training losses keep oscillating without exhibiting a downward trend.  To furthre aid the training of the straightness detection DCNN, we introduce more prior knowledge by initializing the DCNN with the weights only slightly different from the designed corner detection templates in Fig.~\ref{fig:kernels}.  Even so, the training still fails to converge.
The reason for the failures is that the loss function, given the deterministic network architecture, is highly discontinuous near the solution point, very much like in integer programming.  Although the hand crafted network architecture is suitable for the straightness cognition task, its parameters cannot be correctly learnt by the back propagation algorithm of DL.  This exposes, in our view, a serious handicap of DL due to its optimization methodology of the variational calculus.

Similar technical development and conclusions can also be made on the cognitive task of convexity.  As the defining feature of convex vs.\ concave shape classification is whether any of the interior angles exceeds $\pi$, one can design and incorporate into a DCNN a set of convolutional kernels that respond to obtuse angles.  With these hand-crafted kernels the DCNN can be trained to judge convexity and this DCNN solution can be proven to generalize to any object size.

The important message of this section is that the limitations of i.i.d.\ statistical inference for pure data-driven DCNNs can be overcome by using cognitive priors.  These priors can be implemented as predetermined hand-crafted convolutional kernels, which are "innate" to the neural network instead of being nurtured by training data.
This human-enhanced learning approach parallels the learning model of "nurturing the nature" in cognitive psychology.

\section{Conclusions}

This is a study of the limitations of pure data-driven deep learning in cognitive computing.  Despite much success of DL in visual recognition and classification applications, the existing DCNNs were found inept, via extensive experiments similar to those in cognitive psychology, to learn simple spatial concepts that are taken for granted for preschoolers.  They failed most of spatial cognition tasks when the inference was made on instances drawn outside the distribution of the training images.  On the other hand, by using manually crafted features DCNNs can overcome the limitation of i.i.d.\ statistical inference.

\bibliography{spatial}
\bibliographystyle{ieee}

\end{document}